\documentclass[runningheads]{llncs}

\usepackage[utf8]{inputenc}
\usepackage[T1]{fontenc}
\usepackage{microtype}

\usepackage{bm}
\usepackage{graphicx}
\usepackage{amsmath,amsfonts}
\usepackage{booktabs}
\usepackage{multirow}
\usepackage{makecell}
\usepackage{textcomp}
\usepackage{array}
\usepackage{textcomp}
\usepackage{subcaption}
\usepackage{multirow}
\usepackage{float}
\usepackage{xspace}
\usepackage[export]{adjustbox}
\usepackage{nicefrac}
\usepackage{colortbl}
\usepackage{tablefootnote}
\usepackage[referable]{threeparttablex}
\usepackage{comment}

\usepackage[linesnumbered,ruled,vlined]{algorithm2e}
\SetKwRepeat{Do}{do}{while}
\DontPrintSemicolon

\SetCommentSty{commentsty}

\usepackage{mathtools}

\DeclarePairedDelimiterX{\infdivx}[2]{(}{)}{%
	#1\;\delimsize\|\;#2%
}

\makeatletter
\@ifclassloaded{acmart}{}{
\usepackage[usenames,svgnames]{xcolor}
\usepackage{hyperref}
\hypersetup{
    colorlinks = true,
    breaklinks = true,
    bookmarksdepth = section,
    citecolor = DarkGreen,
    linkcolor = DarkRed,
    urlcolor = DarkBlue,
}}
\@ifclassloaded{llncs}{
\usepackage[square,comma,numbers,sort&compress]{natbib}
\bibliographystyle{splncsnat}

}{
\usepackage{amsthm}

\theoremstyle{remark}

}

\makeatother

\DeclareMathOperator*{\argmin}{argmin}

\def\1{\bm{1}}

\DeclareMathAlphabet{\mathsfit}{\encodingdefault}{\sfdefault}{m}{sl}
\SetMathAlphabet{\mathsfit}{bold}{\encodingdefault}{\sfdefault}{bx}{n}

\begin{document}

\title{When Contrastive Learning Meets Active Learning: A Novel Graph Active Learning Paradigm with Self-Supervision}
\titlerunning{A Novel Graph Active Learning Paradigm with Self-Supervision}

\author{Yanqiao Zhu\thanks{These authors have contributed equally to this work.} \and
Weizhi Xu\textsuperscript{\thefootnote} \and
Qiang Liu \and
Shu Wu}

\authorrunning{Y. Zhu et al.}

\institute{Center for Research on Intelligent Perception and Computing\\ Institute of Automation, Chinese Academy of Sciences
\and
School of Artificial Intelligence, University of Chinese Academy of Sciences
\email{\{yanqiao.zhu, weizhi.xu\}@cripac.ia.ac.cn} \\
\email{\{qiang.liu, shu.wu\}@nlpr.ia.ac.cn}
}

\maketitle

\begin{abstract}
This paper studies active learning (AL) on graphs, whose purpose is to discover the most informative nodes to maximize the performance of graph neural networks (GNNs).
Previously, most graph AL methods focus on learning node representations from a carefully selected labeled dataset with large amount of unlabeled data neglected.
Motivated by the success of contrastive learning (CL), we propose a novel paradigm that seamlessly integrates graph AL with CL. While being able to leverage the power of abundant unlabeled data in a self-supervised manner, nodes selected by AL further provide semantic information that can better guide representation learning.
Besides, previous work measures the informativeness of nodes without considering the neighborhood propagation scheme of GNNs, so that noisy nodes may be selected.
We argue that due to the smoothing nature of GNNs, the central nodes from homophilous subgraphs should benefit the model training most. To this end, we present a minimax selection scheme that explicitly harnesses neighborhood information and discover homophilous subgraphs to facilitate active selection.
Comprehensive, confounding-free experiments on five public datasets demonstrate the superiority of our method over state-of-the-arts.

\keywords{Active learning \and contrastive learning \and graph representation learning.}
\end{abstract}

\section{Introduction}
Graph neural networks (GNNs), as a promising means of learning graph representations, attract a lot of research interests \cite{Kipf:2016tc,Velickovic:2018we,Hu:2019vq}.
Most existing GNN models are established in a semi-supervised manner, where limited labeled nodes are given. However, with the same amount of labeled data, the quality of labels strongly affects the model performance \cite{Kipf:2016tc}. Intuitively, a question arises: \emph{how to select high-quality labels to maximize the performance of GNN models?} Active learning (AL), which iteratively selects the most informative samples with the greatest potential to improve the model performance, is widely used to solve this problem.

One of the most critical components in designing an AL algorithm is measuring the informativeness of an instance for labeling.
Existing graph AL methods \cite{Cai:2017wm,Gao:2018wh,Wu:2019wz} design various criteria, which can be roughly categorized into two lines: uncertainty- and representativeness-based strategies.
The former method queries the sample with the least confidence to the model by computing the entropy based on the predicted label distribution, for instance, the samples with a 50\% probability of being positive in binary classification. The latter approach focuses on instances that are representative of the data distribution, e.g., the node closest to the clustering center or with the highest PageRank score.

Though promising progress has been made by previous attempts, these work follows a traditional AL paradigm that learns node representations from an actively selected labeled dataset. Recent surge in self-supervised learning (SSL) suggests that abundant unlabeled data is useful for learning representations \cite{Chen:2020wj,He:2020tu,Caron:2018ba}, which is largely ignored by current AL work.
In this work, we propose to unify graph AL with CL, where we argue the two components could benefit each other. To be specific, on the one hand, CL maximizes the agreement among stochastically augmented versions of the original graph, which empowers AL to leverage the whole dataset without human annotations; on the other hand, AL provides additional semantic information of informative nodes via active selection, which adds positive samples from the same class and further better guides representation learning.


In addition, previous work 
fails to explicitly consider the information propagation scheme of GNNs,
which may end up selecting nodes detrimental to model training. For example, a central node with high entropy and PageRank scores but having dense, noisy inter-class connections may be selected. Labeling such kind of nodes will cause indistinguishable representations, since the embeddings of neighborhood nodes belonging to different communities are smoothed after propagation \cite{Chen:2020cn}.
On the contrary, if nodes in an ego network all share the same label, it is easy for the model to make correct prediction once we have obtained the label for the central node. Therefore, we model the active selection as selecting the central nodes from \emph{homophilous ego networks}, where most nodes are of the same cluster.

\begin{figure}[t]
	\centering
	\includegraphics[width=0.55\linewidth]{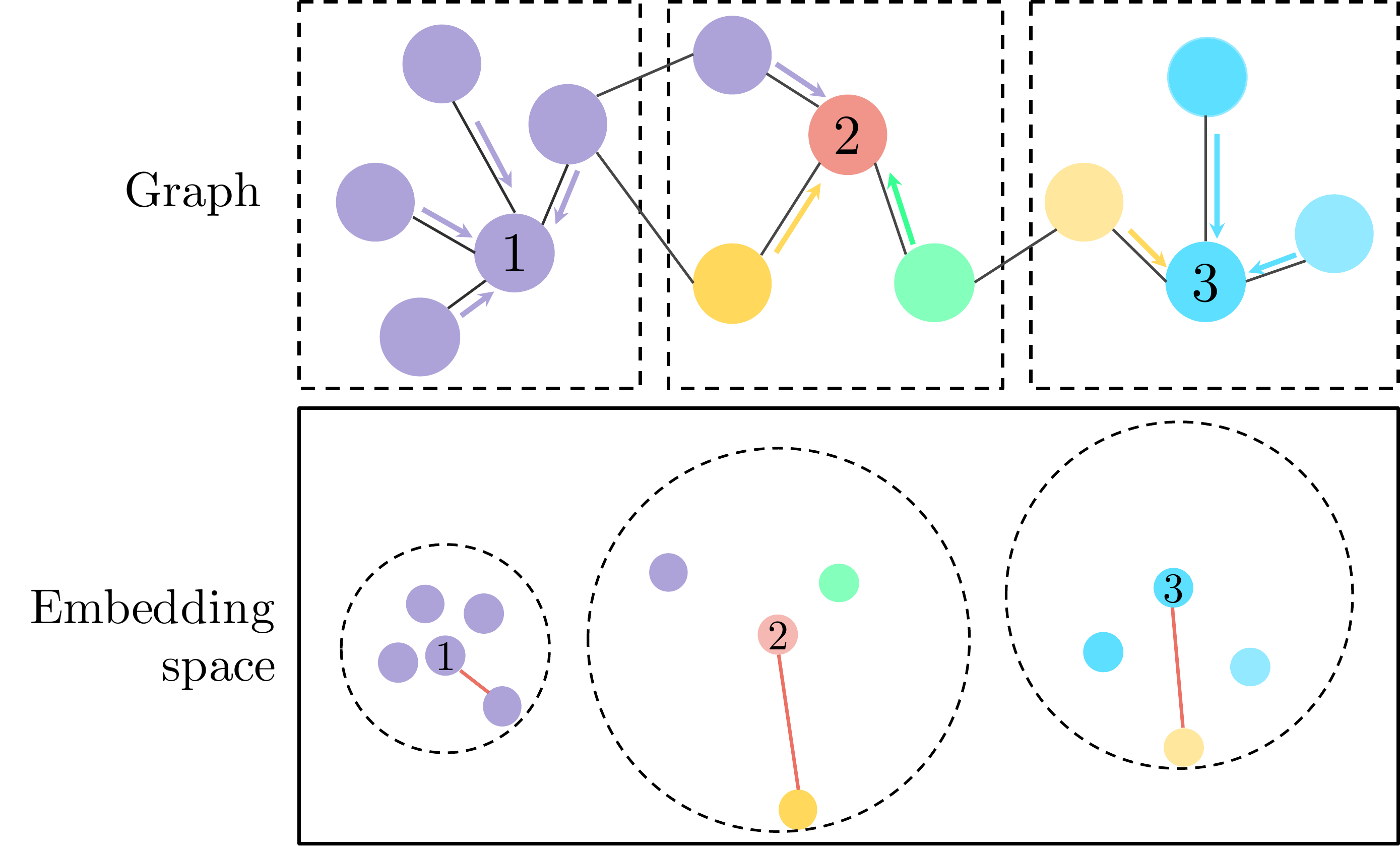}
	\caption{\small Illustrating the proposed minimax selection scheme, including three one-ego networks shown in dashed boxes. Arrows denote neighborhood aggregation; nodes in each subgraph are projected to an embedding space, where the red line represents the furthest distance between the central node and its neighbors. According to our minimax scheme, node 1 will be selected, whose neighboring nodes are densely clustered, indicating high homophily of its subgraph.}
	\label{fig:example}
	\vskip-2em
\end{figure}

However, as most labels remain unavailable during training in AL, we turn to approximating the homophily score of an ego network via similarities of node embedding pairs. This is motivated by the smoothness assumption of semi-supervised learning, which states that instances close to each other in a dense area belong to the same cluster with a high probability \cite{Chapelle:2006vz}.
Accordingly, we introduce a novel minimax active selection scheme. As illustrated in Fig. \ref{fig:example}, we first assign each node with a score defined as the furthest distance in embedding space between itself and its neighbors in its \(k\)-ego network, where \(k\) is the number of GNN layers. Then, we select the node with the minimum score.
By explicitly considering the neighborhood information, the proposed scheme is able to discover homophilous subgraphs in which nodes are the most densely clustered in the embedding space (as shown in the leftmost subgraph in Figure \ref{fig:example}), which benefits the model training.

\begin{figure}[t]
	\centering
	\includegraphics[width=0.9\linewidth]{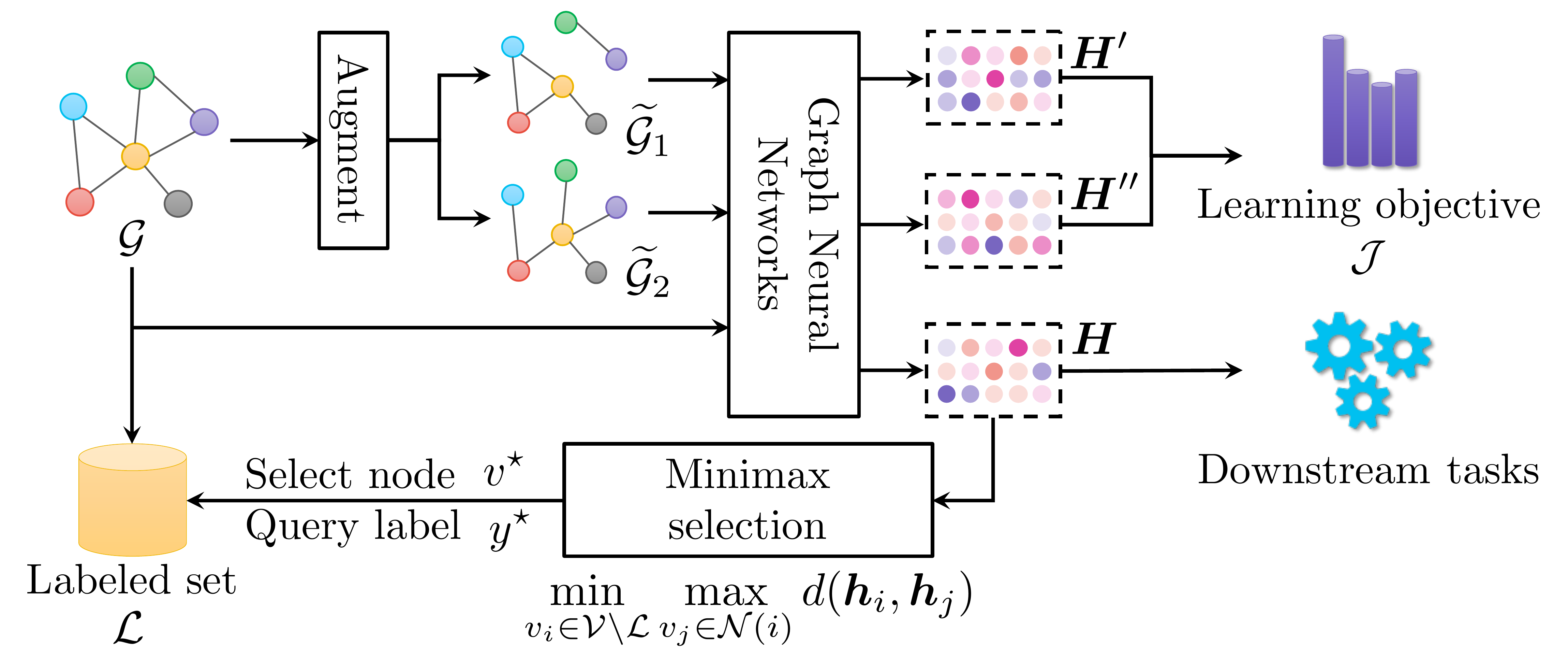}
	\caption{\small The overall framework of our proposed method. In each iteration, we first augment the original graph twice and maximize the congruency between augmented embeddings. Then, we compute the maximum distances between nodes and its neighbors and select the node which minimizes this distance. The model is trained to minimize the unified learning objective \(\mathcal{J}\). The algorithm ends when the number of labeled nodes reaches the budget.}
	\label{fig:model}
	\vskip-1em
\end{figure}

The overall framework is presented in Fig. \ref{fig:model}. In a nutshell, the main contribution of this paper can be summarized as follows.
\begin{itemize}
	\item \textbf{Novel paradigm.} We propose a unified paradigm to empower graph AL with self-supervision. Please kindly note that the CL component is not simply used as a preprocessing step, but is \emph{jointly trained with AL}. The unified \mbox{paradigm is} able to leverage the power of unlabeled dataset; nodes from AL in \mbox{turn provide} additional semantics information to boost classification performance. To the best of our knowledge, this is the first work that has bridged AL with CL.
	\item \textbf{Principled methodology.} Our proposed minimax selection scheme is a more principled approach to GNN-based AL. We explicitly consider the neighborhood aggregation scheme when designing selection criteria and model active selection as selecting the central node from homophilous ego networks. We then propose a novel propagation-aware minimax selection criterion to discover central nodes in homophilous subgraph.
	\item \textbf{Empirical studies.} We have conducted extensive empirical studies without confounding experimental designs on five real-world graph datasets of different scales. The results show the superiority of our method over both traditional and graph-based baselines.
\end{itemize}

\section{The Proposed Method}

In this section, we give the problem definition and notation description. Then, we introduce the proposed paradigm that unifies graph AL with CL, followed by details of the minimax active selection scheme. Lastly, we summarize the overall framework of our proposed method.

\subsection{Preliminaries}

\paragraph{Problem definition.}
Active learning aims to train an accurate model using training data of a limited budget. Specifically, given a large unlabeled data pool $\mathcal{U}^0$, an empty initial labeled set \(\mathcal{L}^0\), and a budget $b$, the model aims to select the top-$b$ informative nodes via the designed selection criteria and train the model with labeled nodes to maximize the performance.

\paragraph{Graph representation learning.}
Let $\mathcal{G}=(\mathcal{V}, \mathcal{E})$ be a graph with $n$ nodes, where $\mathcal{V} = \{v_i\}_{i=1}^n$ is the vertex set and \(\mathcal{E} \subseteq \mathcal{V} \times \mathcal{V}\) is the edge set. We denote the adjacency matrix as \(\bm{A} = \{0,1\}^{n \times n}\) and the node feature matrix as \(\bm{X} \in \mathbb{R}^{n \times m}\), where \(m\) is the dimension of features.
In this paper, following previous work \cite{Cai:2017wm,Gao:2018wh}, we choose the widely-used GCN \cite{Kipf:2016tc} model to learn node representations. Mathematically, the layer-wise propagation rule in GCN can be formulated as
\begin{equation}
	\bm{H}^{(l+1)} = \sigma\left(\widehat{\bm{D}}^{-\frac{1}{2}} \widehat{\bm{A}} \widehat{\bm{D}}^{-\frac{1}{2}} \bm{H}^{(l)} \bm{W}^{(l)}\right),
	\label{eq:gcn}
\end{equation}
where $\widehat{\bm{A}} = \bm{A} + \bm{I}_{n}$ is the adjacency matrix with self-loops and $\widehat{\bm{D}}_{i i}=\sum_{j} \widehat{\bm{A}}_{i j}$. $\bm{H}^{(l)} \in \mathbb{R}^{n \times m_l}$ represents the node embedding matrix in the $l$-th layer, where \(m_l\) is the dimension of the node embedding and \(\bm{H}^{(0)} = \bm{X}\). ${\bm{W}^{(l)}} \in \mathbb{R}^{m \times m_l}$ is a learnable weight matrix of layer \(l\) and $\sigma(\cdot)$ is the nonlinear activation function. We utilize a two-layer GCN model, denoted as \(f(\bm{A}, \bm{X})\). For brevity, we denote \(\bm{H}\) = \(\bm{H}^{(2)}\) as the output representations.

\subsection{Unifying Contrastive Learning and Graph Active learning}
Previous work of graph AL fails to utilize the large amount of unlabeled data during training, which leads to the poor quality of embeddings at the beginning. Motivated by recent progress of self-supervised representation learning, we propose to unify graph AL with CL techniques.

\subsubsection{Graph contrastive learning.}
Graph CL aims at maximizing the agreement between augmented views and promotes the node representations to be invariant to perturbation \cite{Zhu:2020vf,You:2020ut}. To be specific, we first draw two stochastic augmentation functions \(t, t' \sim \mathcal{T}\) from a set of all possible functions \(\mathcal{T}\). Accordingly, we generate two graph views \(\widetilde{\mathcal{G}}_1 = t(\mathcal{G})\) and \(\widetilde{\mathcal{G}}_2 = t'(\mathcal{G})\) and obtain their node embeddings via a shared GNN, denoted by \(\bm{H}' = f(\widetilde{\bm{A}}_1, \widetilde{\bm{X}}_1)\) and \(\bm{H}'' = f(\widetilde{\bm{A}}_2, \widetilde{\bm{X}}_2)\) respectively. Finally, we employ a contrastive objective function \(\mathcal{J}\) to train the GCN model in a self-supervised manner.

\paragraph{Data augmentation.}
Following previous work \cite{You:2020ut,Zhu:2020ui}, we modify both graph structures and node attributes to comprehensively generate graph views.
For topology-level augmentation, we randomly remove edges by sampling probability \(\widetilde{\bm{R}}_{ij}\sim \operatorname{Bern}(1 - p_e)\) for each edge from a Bernoulli distribution, where \(p_e\) denotes the probability of each edge being removed. When there is no edge between two nodes in the original graph, the corresponding entry in \(\widetilde{\bm{R}}\) is set to 0. Then, the resulting corrupted graph structure can be represented as 
\[
\widetilde{\bm{A}} = \bm{A} \circ \widetilde{\bm{R}},
\]
where \(\circ\) denotes element-wise multiplication. Note that other augmentations such as subgraph sampling \cite{Qiu:2020gq} and node dropping \cite{You:2020ut} could be regarded as a special case of edge removing.

For node-level augmentation, we randomly mask partial dimensions of node attributes with zeros. Specifically, we generate a vector \(\widetilde{\bm{m}}\) whose each entry is sampled from another Bernoulli distribution \(\widetilde{\bm{m}}_{ij} \sim \operatorname{Bern}(1-p_n)\) and \(p_n\) is the probability of one dimension being masked. Then, we compute the corrupted attributes as
\[
	\widetilde{\boldsymbol{X}}=\left[\boldsymbol{x}_{1} \circ \widetilde{\boldsymbol{m}} ; \enspace \boldsymbol{x}_{2} \circ \widetilde{\boldsymbol{m}} ; \enspace \cdots ; \enspace \boldsymbol{x}_{n} \circ \widetilde{\boldsymbol{m}}\right],
\]
where \([\cdot;\cdot]\) denotes the concatenation operator and \(\bm{x}_i\) is the \(i\)-th row of \(\bm{X}\).

\paragraph{Contrastive objective.}
Thereafter, we employ a contrastive objective function that learns the node embeddings in a self-supervised manner by pulling embeddings of the same node together and pushing that of different nodes apart.
Specifically, for an anchor \(\bm{h}_i'\), we first define its pairwise contrastive objective, which takes in the form of NT-Xent loss \cite{Chen:2020wj}:
\begin{equation}
	\ell(\bm{h}_i', \bm{h}_i'') = - \log \frac{\theta(\bm{h}_i', \bm{h}_i'')}{\theta(\bm{h}_i', \bm{h}_i'') + \sum_{j \neq i} \left[ \theta(\bm{h}_i', \bm{h}_j') + \theta(\bm{h}_i', \bm{h}_j'') \right] }.
	\label{eq:pairwise-objective}
\end{equation}
The critic function \(\theta(\cdot, \cdot)\) computes the cosine similarity between two embeddings scaled by a temperature scalar \(\tau \in \mathbb{R}^+\):
\[
	\theta(\bm{h}', \bm{h}'') = \exp\left( \frac{g(\bm{h}')^\top g(\bm{h}'')}{\tau \cdot \|g(\bm{h}')\| \|g(\bm{h}'')\|} \right),
\]
where \(g(\cdot)\) is a non-linear projection function, e.g., a multilayer perception.
Since the two views are symmetric, the final contrastive objective is defined as the average over all nodes:
\begin{equation}
	\mathcal{J}_\text{CL} = \frac{1}{2n} \sum_{v_i \in \mathcal{V}} \left[\ell(\bm{h}_i', \bm{h}_i'') + \ell(\bm{h}_i'', \bm{h}_i') \right].
	\label{eq:contrastive-objective}
\end{equation}

\subsubsection{Incorporating semantic information from active learning.}
In a pure unsupervised learning setting, we can see that for an anchor \(\bm{h}_i'\), the computation in Eq. (\ref{eq:pairwise-objective}) involves only one positive sample, and \(2(n - 1)\) negative samples.
However, in our graph AL scenarios, we need to consider semantic information of nodes selected by AL algorithms. Due to the presence of labels, we further encourage the GNN encoder to include positive samples belonging to the same class, resulting in a more aligned and compact embedding space.

To generalize CL to AL, we modify the contrastive objective in Eq. (\ref{eq:pairwise-objective}) as
\begin{equation}
	\small
	\ell(\bm{h}_i', \bm{h}_i'') = - \log \frac{ \theta(\bm{h}_i', \bm{h}_i'') + \lambda \cdot \sum_{\bm{h}_p \in \mathcal{P}(i)} \theta(\bm{h}_i', \bm{h}_p)}{\theta(\bm{h}_i', \bm{h}_i'') + \lambda \cdot \sum_{\bm{h}_p \in \mathcal{P}(i)} \theta(\bm{h}_i', \bm{h}_p) + \sum_{j \neq i} \left[ \theta(\bm{h}_i', \bm{h}_j') + \theta(\bm{h}_i', \bm{h}_j'') \right] }.
	\label{eq:active-pairwise-objective}
\end{equation}
Here, besides \(\bm{h}_i''\) we further introduce a positive embedding set \(\mathcal{P}(i)\) consisting of embeddings whose label is the same with node \(v_i\)'s. In particular, to closely examine the contribution of these supervised positives, we further introduce a hyperparameter \(\lambda \in \mathbb{R}^+\) to balance the two terms.
Similar to Eq. (\ref{eq:contrastive-objective}), we define our final objective as summation over all nodes:
\begin{equation}
	\mathcal{J} = \frac{1}{2n} \sum_{v_i \in \mathcal{V}} \frac{1}{|\mathcal{P}(i)| + 1} \left[\ell'(\bm{h}_i', \bm{h}_i'') + \ell'(\bm{h}_i'', \bm{h}_i') \right].
	\label{eq:active-contrastive-objective}
\end{equation}

\subsection{The Minimax Active Selection Scheme}

Having introduced the active-contrastive learning framework, we further describe our proposed minimax selection scheme in detail.
Consider that the neighborhood aggregation scheme of GNNs could be regarded as Laplacian smoothing, which makes representations of neighboring nodes similar \cite{Li:2018wc,Chen:2020vu}.
Ideally, if nodes in an ego network share the same label, once we label the central node in that network, the remaining nodes will get the correct prediction due to smoothed embeddings \cite{Chen:2020cn}.
We thus argue that the active node selection could be formulated as selecting the central node from homophilous \(k\)-ego networks, where the homophily \mbox{score is} defined by the number of neighbors belonging to the same class of the \mbox{center node}.

However, labels of unselected nodes remain unavailable during training. We then turn to select the nodes close to their neighbors in the embedding space, since they are more likely to fall into the same community according to the smoothness assumption in semi-supervised learning \cite{Chapelle:2006vz}.
Thereafter, we propose a minimax-based selection scheme to measure the neighborhood similarity and select the central nodes from the most homophilous subgraphs.
Specifically, we first discover the furthest neighbor of each node in the embedding space. Among these nodes, we select the node with the shortest distance such that it has the most probability to constitute a homophilous subgraph. Formally, the node \(v^\star\) we select according to the minimax selection scheme is
\begin{equation}
	v^\star = \argmin_{v_i \in \mathcal{V} \setminus \mathcal{L}} \max_{v_j \in \mathcal{N}(i)}{d(\bm{h}_i, \bm{h}_j)},
	\label{eq:minimax}
\end{equation}   
where \(d(\bm{h}_i, \bm{h}_j) = \| \bm{h}_i - \bm{h}_j \|_2^2\) denotes the Euclidean distance between two node embeddings \(\bm{h}_i\) and \(\bm{h}_j\), \(\mathcal{N}(i)\) denotes the \(k\)-hop neighbor set of node \(i\), and \(\mathcal{L}\) denotes the labeled set.  
Once we have obtained the label of \(v^\star\), we add it to the labeled set and update the positive embedding set \(\mathcal{P}(\star)\) accordingly.

\subsection{The Overall Framework}

Finally, we summarize the proposed method, as shown in Fig. \ref{fig:model} and Algorithm \ref{algo:algorithm}.
At each iteration \(c\), we first generate two different augmented graph views.
Then, we feed two graph views \(\widetilde{G}_1\) and \(\widetilde{G}_2\) into GNNs to compute their node embeddings. After that, along with an enlarged positive set \(\mathcal{P}\) resulting from the labeled dataset \(\mathcal{L}^c\), we train the model using a unified objective function \(\mathcal{J}\).
After training, we calculate the distance between each node and its \(k\)-hop neighbors based on their embeddings and select one node \(v^\star\) via the proposed minimax scheme to query its label.
Finally, we obtain an updated unlabeled set \(\mathcal{U}^{c+1} = \mathcal{U}^{c} \backslash {\{v^\star\}}\) and a labeled set \(\mathcal{L}^{c+1}=\mathcal{L}^{c} \cup {\{(v^\star, y^\star)\}}\).
We repeat the above steps until the size of labeled set reaches the labeling budget \(b\).

\begin{algorithm}[h]
	\DontPrintSemicolon\SetNoFillComment
	\caption{The proposed method}
	\label{algo:algorithm}
	\SetKw{Continue}{continue}
	\KwData{Graph \(\mathcal{G}\), budget \(b\), GNN model \(f\)}
	Initialize the unlabeled set \(\mathcal{U}^0 = \mathcal{V}\) and the labeled set \(\mathcal{L}^0 = \emptyset\)\;
	\For{\(c = 0\) to \(b\)}{
		Construct two graph views \(\widetilde{G}_1\) and \(\widetilde{G}_2\), and obtain embeddings \(\bm{H}'\) and \(\bm{H}''\)\;
		Train the model \(f\) to minimize the learning objective \(\mathcal{J}\) as in Eq. (\ref{eq:active-contrastive-objective}) \;
		Calculate the distance between each node \(v_i\) and its neighbors \(v_j \in \mathcal{N}{(i)}\)\;
		Select node \(v^\star\) according to Eq. (\ref{eq:minimax}) and query its label \(y^\star\)\;
		Update \(\mathcal{P}(\star)\) with node embeddings having the same label as \(y^\star\) \;
		\(\mathcal{L}^{t+1} = \mathcal{L}^t\) $\cup$ \(\{(v^\star, y^\star)\}\)\;
		\(\mathcal{U}^{t+1} = \mathcal{U}^{t}  \backslash  {\{v^\star\}}\)\;
	}
	\Return labeled set \(\mathcal{L}\)\;
\end{algorithm}

\section{Experiments}
The experiments presented in this section aim to answer the following questions.
\begin{itemize}
	\item \textbf{Q1}. How does CL benefits the AL performance?
	\item \textbf{Q2}. How is the proposed minimax selection criterion compared with existing graph AL approaches?
	\item \textbf{Q3}. How do key hyperparameters affect the model performance?
\end{itemize}
\subsection{Experimental Setup}

\paragraph{Datasets.}
Following previous work \cite{Cai:2017wm,Gao:2018wh}, we use three citation network datasets: Cora, Citeseer and Pubmed. Each dataset contains a graph, where nodes represent articles and edges represent citation relationship, and the node features are sparse bag-of-words feature vectors.
In addition, considering the three datasets are of small scales, we include two larger datasets: Computer and Photo, where nodes indicate items and edges indicate co-purchase relationship from the Amazon dataset
Detailed statistics of datasets is summarized in Table \ref{tab:statistics}. The datasets we use in experiments are all publicly available.

\begin{table}[t]
	\centering
	\caption{\small Statistics of five datasets used in experiments.}
	\vskip0.5em
	\label{tab:statistics}
	\begin{tabular}{cccccc}
		\toprule
		Dataset & \#Nodes & \#Edges & \#Classes & \#Features \\
		\midrule
		Cora     & 2,708  & 5,429  & 7       & 1,433   \\
		Citeseer & 3,327  & 4,732  & 6       & 3,703   \\ 
		Pubmed   & 19,717 & 44,338 & 3       & 500     \\
		Computer & 13,752 & 245,861 & 10 & 767 \\
		Photo & 7,650 & 119,081 & 8 & 745 \\
		\bottomrule
	\end{tabular}
\end{table}

\paragraph{Baselines.}
To evaluate the performance of our proposed approach, we compare it with six representative baselines, including three traditional AL methods (Random, Degree, and Entropy), and three graph-based AL methods (AGE, ANRMAB, and FeatProp).
\begin{itemize}
	\item \textbf{Random}. All training data are randomly selected.
	\item \textbf{Degree}. We select the node with the largest degree in each iteration.
	\item \textbf{Entropy}. We select the node that has the maximum entropy of the predicted label distribution in each iteration.
	\item \textbf{AGE} \cite{Cai:2017wm}. AGE designs three selection criteria, calculating uncertainty via entropy of the predicted label distribution, measuring node centrality via the PageRank algorithm, and obtaining node density by calculating distances between nodes and cluster centers.
	\item \textbf{ANRMAB} \cite{Gao:2018wh}. ANRMAB uses the same criteria as AGE and further applies a multi-armed bandit mechanism to adaptively adjust the importance of these criteria.
	\item \textbf{FeatProp} \cite{Wu:2019wz}. FeatProp applies the K-medoids algorithm to cluster nodes based on node features and selects the central node for labeling.
\end{itemize}

\paragraph{Implementation details.}
For fair comparison, we closely follow the experimental setting in previous work \cite{Cai:2017wm,Gao:2018wh}.
For each dataset, we randomly sample 500 nodes for validation and 1,000 nodes for test to ensure that the performance variance is due to AL strategies.

We train a two-layer GCN model with 128 hidden units for a maximum of 200 epochs using the Adam optimizer \cite{Kingma:2015us} with a learning rate of 0.001. We set \(\lambda = 1.0\) after grid search on the validation set. Similar to existing CL work, after training the model, we train a logistic regression model on the learned node embeddings. We repeat training and test for 10 times on 10 different validation sets and report the averaged performance.

The proposed method is implemented using PyTorch 1.5.1 \cite{Paszke:2019vf} and PyTorch Geometric 1.6 \cite{Fey:2019wv}. All experiments are conducted on a Linux server equipped with four NVIDIA Tesla V100S GPUs (each with 32GB memory) and 12 Intel Xeon Silver 4214 CPUs.

\subsection{Overall Performance (Q1--Q2)}

\paragraph{Confounding-free experimental configurations.}
For simplicity, we denote the number of classes as \(C\) hereafter.
First of all, to demonstrate the effectiveness of the unified graph AL paradigm, we conduct comparative experiments with a budget of \(20C\) nodes, with and without CL components.
Specifically, we utilize a widely-used CL framework GRACE \cite{Zhu:2020vf} \emph{on top of all baselines} such that the comparison can be made on the same basis.
As a comparison, following previous work AGE \cite{Cai:2017wm} and ANRMAB \cite{Gao:2018wh}, we also report model performance for all baselines without CL components, but \emph{with an initial dataset} that consists of \(4C\) labeled nodes.

In addition, to verify the effectiveness of our proposed minimax selection criteria, we further conduct experiments under smaller budget settings (30 and 60 nodes), closely following FeatProp \cite{Wu:2019ke}. In this case, all baselines are employed with CL components and without initial pools.

\subsubsection{Contrastive learning v.s. initial labeled pools (Q1).}

\begin{figure}
	\centering
	\includegraphics[width=\linewidth]{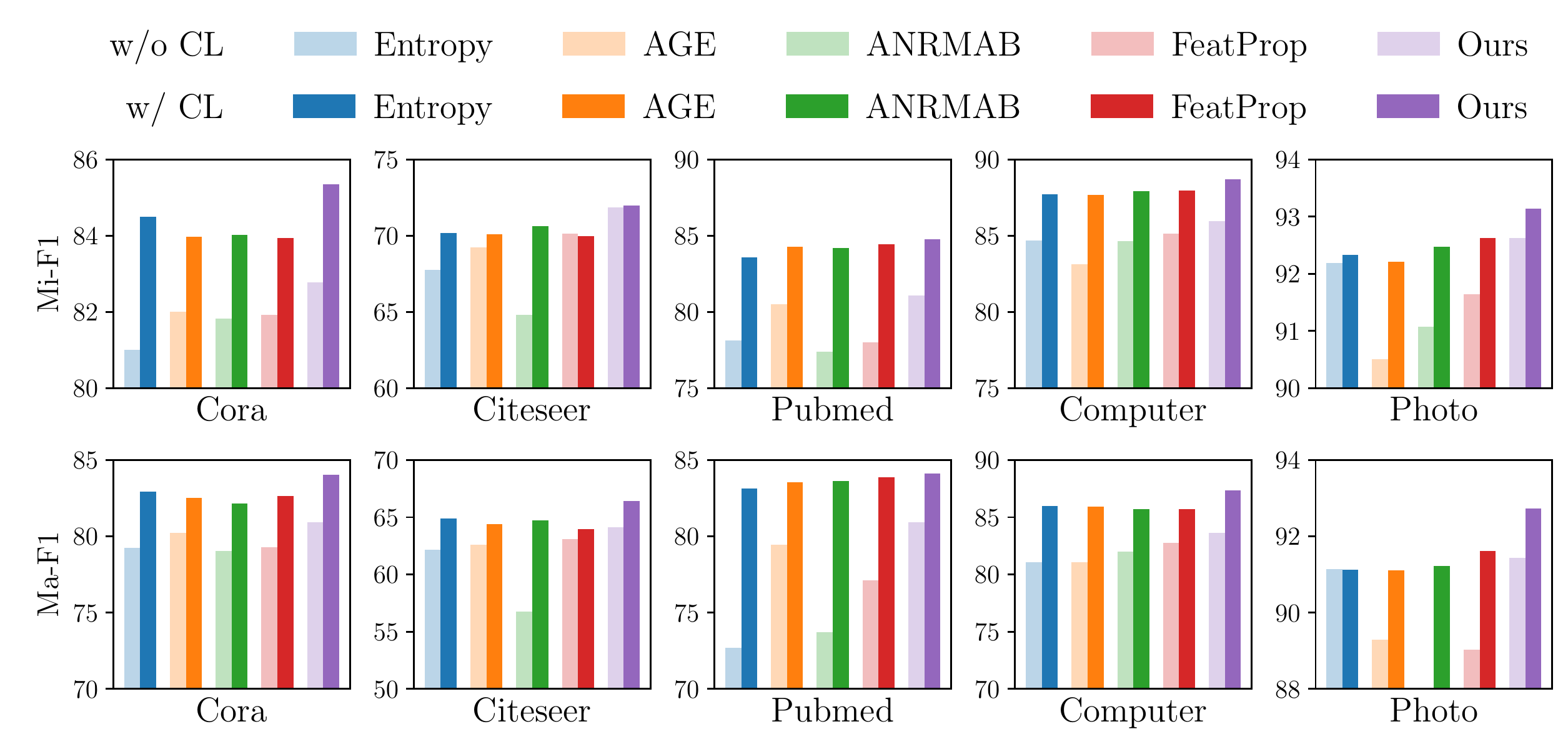}
	\caption{\small Performance comparison of our proposed unified AL paradigm for baselines with and without CL components in terms of Micro-F1 (Mi-F1) and Macro-F1 (Ma-F1) scores (\%).}
	\label{fig:performance}
\end{figure}

The performance of the comparative study is summarized in Fig. \ref{fig:performance}.
It is obvious that \emph{all AL approaches} trained with CL components show superior performance to their counterparts, which are initialized with a labeled pool. This indicates that, on the one hand, our proposed framework is able to effectively leverage the power of abundant unlabeled data via self-supervised learning. It benefits AL selection strategies by providing a well-aligned embedding space. On the other hand, as nodes selected via AL strategies offer rich semantic information, CL could in turn benefit from such valuable information to find more positive samples, resulting in a more robust embedding space.
In summary, the performance demonstrates the effectiveness of our proposed unified paradigm, which boosts performance of AL methods with no initial pool given.

\subsubsection{The minimax selection scheme v.s. baselines (Q2).}

We further experiment with smaller budget configurations to comprehensively study the proposed minimax selection criterion, where the results are presented in Table \ref{tab:main-exp}.
Again, for fair comparison, we employ the same CL components on all node selection baselines, except for Random and Degree since they do not select nodes based on embeddings.
From the table, we can see that our proposed minimax selection scheme outperforms all traditional and graph-based baselines on five datasets by significant margins, which demonstrates the superiority of our method.

Furthermore, there are several interesting findings regarding the results.
\begin{itemize}
	\item The performance of traditional AL baselines is relatively worse than that of graph-based methods, indicating that both structural and attributive information is necessary for AL on graph-structured data. What stands out from the table is that selecting nodes merely according to node degrees results in severe performance downgrading on the Computers dataset. We suspect that it is primarily due to the large scale and relatively low homophily of the dataset (c.f., Table \ref{tab:homophily}), where high-degree nodes may be connected with many inter-class edges; selecting such nodes brings no benefits for accurate prediction.
	\item Graph-based baselines achieve inferior performance to that of ours. Once again, this suggests that selecting nodes according to uncertainty and representativeness measures without taking the GNN propagation scheme into consideration will lead to suboptimal performance.
	\item We summarize the average homophily score (i.e., the ratio of intra-class edges to all edges) of all one-ego networks of the selected nodes. As illustrated in Table \ref{tab:homophily}, we observe that the homophily scores of subgraphs induced by the selected nodes are much higher than the average score of the original graph on all datasets, which verifies that our proposed method is able to select central nodes from homophilous subgraphs.

\end{itemize}

\begin{table}
	\centering
    \caption{\small Performance of node classification in terms of Micro-F1 (Mi-F1) and Macro-F1 (Ma-F1) scores (\%) with different budgets: 30, 60, and 20\(C\). For fair comparison, all models are provided with no initial pools and are trained with the same CL component. Note that 20\(C\) is equal to 60 on the Pubmed dataset.}
	\vskip0.5em
	\label{tab:main-exp}
	\begin{subtable}[h]{\linewidth}
	\centering
	\caption{\small Experiments with default budget configuration \(b = 20C\)}
    \begin{tabular}{ccccccccc}
    \toprule
    Dataset & Metric & Random & Degree & Entropy & AGE & ANRMAB & FeatProp & Ours \\
    \midrule
    \multirow{1.5}[2]{*}{Cora} & Mi-F1 & 78.31 & 80.68 & 84.50 & 83.97 & 84.02 & 83.94 & \textbf{85.35} \\
          &      Ma-F1 & 79.11 & 79.42 & 82.92 & 82.52 & 82.14 & 82.63 & \textbf{84.02} \\
    \midrule
    \multirow{1.5}[2]{*}{Citeseer} & Mi-F1 & 67.80 & 67.21 & 70.18 & 70.08 & 70.63 & 69.95 & \textbf{71.98} \\
          &      Ma-F1 & 62.91 & 59.93 & 64.91 & 64.40 & 64.74 & 63.97 & \textbf{66.44} \\
    \midrule
    \multirow{1.5}[2]{*}{Pubmed} & Mi-F1 & 77.14 & 77.61 & 83.98 & 84.26 & 84.21 & 84.45 & \textbf{84.76} \\
          &      Ma-F1 & 76.98 & 76.92 & 83.13 & 83.53 & 83.62 & 83.88 & \textbf{84.11} \\
    \midrule
    \multirow{1.5}[2]{*}{Computer} & Mi-F1 & 84.21 & 71.75 & 87.71 & 87.67 & 87.91 & 87.94 & \textbf{88.69} \\
          &    Ma-F1 & 77.83 & 44.09 & 85.99 & 85.92 & 85.69 & 85.73 & \textbf{87.33} \\
    \midrule
    \multirow{1.5}[2]{*}{Photo} & Mi-F1 & 91.97 & 86.19 & 92.33 & 92.21 & 92.47 & 92.63 & \textbf{93.14} \\
          &  Ma-F1 & 91.05 & 84.89 & 91.12 & 91.11 & 91.23 & 91.62 & \textbf{92.73} \\
    \bottomrule
    \end{tabular}
	\end{subtable}
	
	\begin{subtable}[h]{\linewidth}
	\caption{\small Experiments with a smaller budget configuration (\(b = 30\) and \(b = 60\))}
	\resizebox{\linewidth}{!}{
    \begin{tabular}{cccccccccc}
    \toprule
    Dataset & Budget & Metric & Random & Degree & Entropy & AGE & ANRMAB & FeatProp & Ours \\
    \midrule
    \multirow{4}[2]{*}{Cora} & \multirow{2}[1]{*}{30} & Mi-F1 & 68.23 & 69.81 & 83.44 & 83.74 & 83.60 & 83.64 & \textbf{84.13} \\
          &       & Ma-F1 & 67.12 & 69.24 & 81.77 & 81.88 & 81.82 & 82.24 & \textbf{83.56} \\ \cmidrule{2-10}
          & \multirow{2}[0]{*}{60} & Mi-F1 & 77.22 & 75.58 & 83.75 & 83.86 & 83.68 & 83.86 & \textbf{84.69} \\
          &       & Ma-F1 & 76.81 & 74.48 & 82.20 & 82.32 & 81.92 & 82.26 & \textbf{83.87} \\
    \midrule
    \multirow{4}[2]{*}{Citeseer} & \multirow{2}[1]{*}{30} & Mi-F1 & 55.86 & 56.21 & 70.18 & 69.96 & 69.51 & 69.79 & \textbf{70.95} \\
          &       & Ma-F1 & 51.33 & 51.66 & 63.82 & 64.02 & 63.21 & 63.84 & \textbf{65.89} \\ \cmidrule{2-10}
          & \multirow{2}[0]{*}{60} & Mi-F1 & 62.27 & 64.06 & 70.68 & 70.12 & 69.74 & 70.22 & \textbf{71.91} \\
          &       & Ma-F1 & 60.18 & 62.23 & 64.31 & 65.18 & 63.92 & 64.73 & \textbf{66.03} \\
    \midrule
    \multirow{4}[2]{*}{Pubmed} & \multirow{2}[1]{*}{30} & Mi-F1 & 72.91 & 73.32 & 84.10 & 84.16 & 84.12 & 84.19 & \textbf{84.45} \\
          &       & Ma-F1 & 70.21 & 72.72 & 83.59 & 83.41 & 83.67 & 83.63 & \textbf{83.98} \\ \cmidrule{2-10}
          & \multirow{2}[0]{*}{60} & Mi-F1 & 77.14 & 77.61 & 83.98 & 84.26 & 84.21 & 84.45 & \textbf{84.76} \\
          &       & Ma-F1 & 76.98 & 76.92 & 83.13 & 83.53 & 83.62 & 83.88 & \textbf{84.11} \\
    \midrule
    \multirow{4}[2]{*}{Computer} & \multirow{2}[1]{*}{30} & Mi-F1 & 72.55 & 53.83 & 87.23 & 87.16 & 87.31 & 87.24 & \textbf{87.90} \\
          &       & Ma-F1 & 68.49 & 40.16 & 85.29 & 85.45 & 85.55 & 84.97 & \textbf{86.48} \\ \cmidrule{2-10}
          & \multirow{2}[0]{*}{60} & Mi-F1 & 79.92 & 56.63 & 87.54 & 87.63 & 87.86 & 87.57 & \textbf{88.33} \\
          &       & Ma-F1 & 73.94 & 42.28 & 85.61 & 85.75 & 85.94 & 85.18 & \textbf{86.72} \\
    \midrule
    \multirow{4}[2]{*}{Photo} & \multirow{2}[1]{*}{30} & Mi-F1 & 88.72 & 42.19 & 92.26 & 92.28 & 92.12 & 92.14 & \textbf{92.45} \\
          &       & Ma-F1 & 88.51 & 51.92 & 90.97 & 91.07 & 91.08 & 90.82 & \textbf{91.52} \\ \cmidrule{2-10}
          & \multirow{2}[0]{*}{60} & Mi-F1 & 92.56 & 46.93 & 92.19 & 92.11 & 92.33 & 92.37 & \textbf{92.59} \\
          &       & Ma-F1 & 91.53 & 57.88 & 91.03 & 91.17 & 91.08 & 91.16 & \textbf{91.58} \\
    \bottomrule
    \end{tabular}
    }
	\end{subtable}

\end{table}

\begin{table}
	\centering
	\caption{\small The average homophily scores of the original graph (Original) and the subgraphs selected by ours (Selected), with a budget \(b = 20C\). The relative improvement is shown in the last row.}
	\vskip0.5em
	\label{tab:homophily}
	\begin{tabular}{cccccc}
		\toprule
		Graph & Cora & Citeseer & Pubmed & Computer & Photo \\
		\midrule
		Original     & 0.810  & 0.736  & 0.802  & 0.778 & 0.828  \\
		Selected & 0.921 & 0.816 & 0.872 & 0.858 & 0.929  \\ 
		\midrule
		Improv. & 13.7\% & 10.9\% & 8.73\% & 10.3\% & 12.2\% \\
		\bottomrule
	\end{tabular}
\end{table}

\subsection{Sensitivity Analysis (RQ3)}
In this section, we conduct sensitivity analysis of two key hyper-parameters in our proposed approach. While we alter one parameter, other experimental configurations remain unchanged.

\begin{figure}
	\centering
	\includegraphics[width=\linewidth]{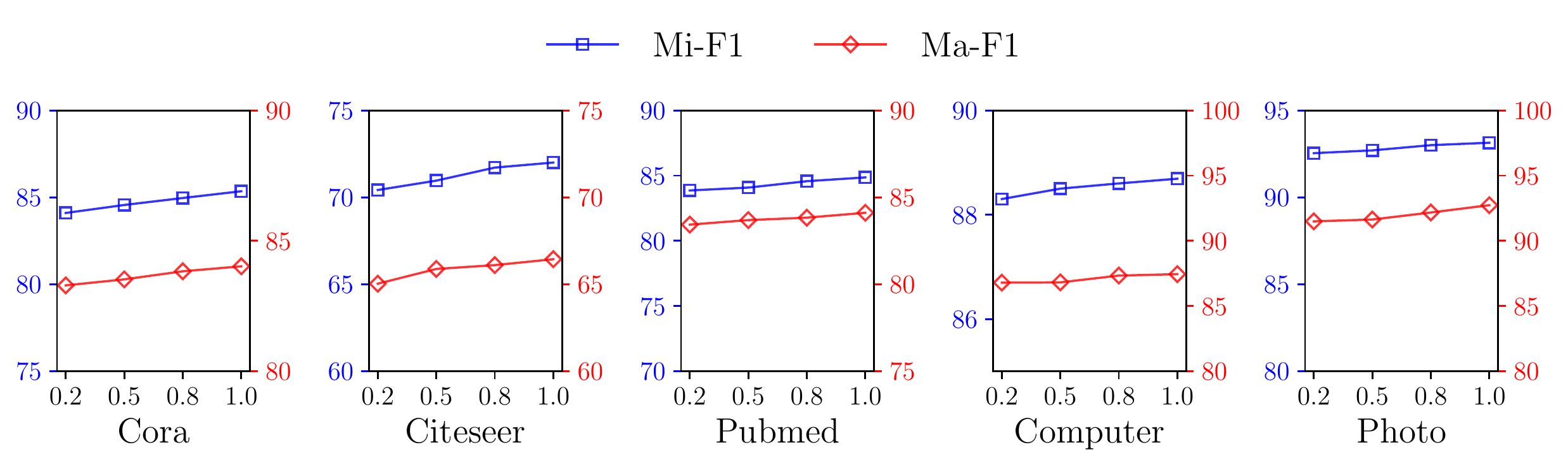}
	\caption{\small Node classification performance with varied \(\lambda\)'s in terms of Mi-F1 and Ma-F1 (\%), with a budget \(b = 20C\) nodes and without an initial pool.}
	\label{fig:lambda-value}
\end{figure}

Firstly, we conduct experiments with different values (0.2, 0.5, 0.8, 1.0) of \(\lambda\), which adjusts the ratio of contribution of the supervised component in the unified loss. From Fig. \ref{fig:lambda-value}, we can observe that the performance improves steadily when \(\lambda\) increases, demonstrating that the supervised information provided by our AL algorithm plays a positive role in model training. We suspect that the main reason behind the performance gain is the rich semantic information involved in labeled nodes promotes CL to pull more positive samples together so that representations are more distinguishable. 

Secondly, we analyze the impact of neighborhoods considered in our minimax selection scheme. We change \(k\) from one- to two-hop neighbors, since deep GCNs will lead to oversmoothing \cite{Li:2018wc}. From Table \ref{tab:k-value}, we note that models using one- and two-hop neighborhood information exhibits negligible difference of performance. We also observe that Ours--1 slightly outperforms Ours--2 on all datasets, which may indicate that some noisy neighboring nodes are included when considering two-hop neighborhoods. Furthermore, since computing distance between two-hop neighborhood pairs introduces extra computational burden, in our experiments, we simply choose to include only one-hop neighbors to speed up computation.

\begin{table}[t]
	\centering
	\caption{\small Performance comparison of models considering one-hop neighborhood (Ours--1) and two-hop neighborhood (Ours--2) in terms of Mi-F1 (\%), with a budget \(b = 20C\) nodes and without an initial pool.}
	\label{tab:k-value}
	\vskip0.5em
	\begin{tabular}{cccccc}
		\toprule
		Model & Cora & Citeseer & Pubmed & Computer & Photo \\
		\midrule
		Ours--1 & 85.35  & 71.98  & 84.76  & 88.69 & 93.14  \\
		Ours--2 & 85.26 & 71.82 & 84.74 & 88.60 & 92.94  \\ 
		\bottomrule
	\end{tabular}
\end{table}

\section{Related Work}
In this section, we briefly review related work of graph contrastive learning. Then, we review prior arts on active learning for both Euclidean data and graphs.

\subsection{Graph Contrastive Learning}
Contrastive learning (CL) is a specific branch of self-supervised learning, which targets at obtaining robust and discriminative representations by contrasting positive and negative instances. In the domain of computer vision, many representative CL methods including DIM \cite{Bachman:2019wp,Hjelm:2019uk}, SimCLR \cite{Chen:2020wj}, and SimSiam \cite{He:2020tu}, achieve superior performance in visual representation learning. Generally, most CL work differs from each other in terms of the generation of negative samples (e.g., color jitter, random flip, cropping, resizing, rotation) and contrastive objectives.

Recently, CL on the graph-structure data has attracted lots of research interest. \citet{Velickovic:2019tu} first design a CL method for GNNs, where they introduce an objective function measuring the Mutual Information (MI) between global graph embeddings and local node embeddings. Similarly, InfoGraph \cite{Sun:2020vi} generalizes DGI to graph-level tasks. Thereafter, MVGRL \cite{Hassani:2020un} performs node diffusion and contrasts node embeddings on corrupted graphs. Follow-up work GraphCL \cite{You:2020ut} and GRACE \cite{Zhu:2020vf} propose a node-level contrastive objective to simplify previous work. Recent work GCA \cite{Zhu:2021wh} proposes stronger augmentation schemes on graphs; GCC \cite{Qiu:2020gq} firstly introduces a pre-training framework for contrastive graph representation learning.

\subsection{Active Learning on Euclidean Data}
Different active learning algorithms propose various strategies to select the most informative instances from a large pool of unlabeled data. Previous approaches can be roughly grouped into three categories \cite{Settles:2009vo}: uncertainty-based, performance-based, and representativeness-based methods.

For the methods falling into the first category, \citet{Settles:2008tf} propose the uncertainty sampling, which calculates uncertainty score based on cross-entropy over the label distribution. \citet{Bilgic:2010uo} introduce a vote mechanism to select the instance who receives the most disagreement votes from the models.
Regarding performance-based algorithms, \citet{Guo:2007vp,Schein:2007wd} propose criteria directly related to the model performance including prediction error and variance reduction.
The last group of methods focus on discovering the instance that is representative of the data distribution. \citet{Sener:2018we} regard the sampling process as a coreset problem, in which the representations of the last layer in deep neural networks are used for constructing the coreset. However, these methods can not be directly performed on graph-structural data, since they are all designed for independent and identical distributed (i.i.d) data and do not consider rich structural information.

\subsection{Active Graph Representation Learning}
Recently, there is a growing interest of AL on graphs. AGE \cite{Cai:2017wm} calculates the informative score by combining three designed criteria. For uncertainty, they compute entropy on the predicted label distribution. For representativeness, they measure the distance between a node and its cluster center and obtain the centrality via the PageRank algorithm \cite{Page:1997qy}. Finally, they combine these criteria with the time-sensitive parameters linearly.
ANRMAB \cite{Gao:2018wh} uses the same selection criteria as AGE and further introduces a multi-armed bandit algorithm to adaptively decide weights of these three criteria. However, these methods ignore the essence of propagation scheme, which may lead the algorithm to select the node with too many inter-class neighbors. \citet{Hu:2020tl} propose to learn a transferable AL policy that decreases the cost of retraining. Besides, FeatProp \cite{Wu:2019wz}, a clustering-based method, calculates distances between every pair of nodes based on the fixed raw node features. Then, it applies the K-Medoids algorithm to clustering nodes by utilizing the distances. Our method also computes node distances, but we model the AL selection from the perspective of neighborhood propagation and explicitly utilize the local neighborhood information, which is fundamentally different from FeatProp.

\section{Conclusion}
In this paper, we have proposed a novel graph active learning framework, which integrates CL and AL seamlessly to leverage the power of abundant unlabeled data. Besides, consider that previous work neglects the neighborhood information in graph AL selection strategies, we propose that the active selection problem can be casted as selecting central nodes in homophilous subgraphs. Then, we design a novel minimax selection scheme that takes into account the neighborhood information. Moreover, comprehensive, confounding-free experiments have been conducted to verify the effectiveness of our proposed framework and the selection strategy, which show that our method outperforms all baselines by significant margins.

\bibliography{ecmlpkdd2021}

\end{document}